\pdfoutput=1
\documentclass[11pt]{article}

\usepackage[dvipsnames]{xcolor}

\usepackage{acl}

\usepackage{times}
\usepackage{latexsym}
\usepackage{ulem}

\usepackage[T1]{fontenc}
\usepackage[utf8]{inputenc}

\usepackage{microtype}

\usepackage{inconsolata}

\usepackage{enumitem}
\usepackage{array}
\usepackage{tabularx}
\usepackage{dashrule}
\usepackage{hyperref}

\usepackage{xspace}
\usepackage{graphicx}

\usepackage{todonotes}

\definecolor{forestgreen}{rgb}{0.13, 0.55, 0.13}
\definecolor{flame}{rgb}{0.89, 0.35, 0.13}
\definecolor{startblue}{RGB}{82, 151, 248}

\newcommand{\model}{\textsc{RENarGen}\xspace}

\usepackage{booktabs}

\title{Returning to the Start: Generating Narratives with Related Endpoints}

\author{Anneliese Brei
        \hspace{.4cm} 
        Chao Zhao \hspace{.4cm} 
        Snigdha Chaturvedi \\
        Department of Computer Science, UNC Chapel Hill\\
        \texttt{\{abrei, zhaochao, snigdha\}@cs.unc.edu}}

\begin{document}
\maketitle
\begin{abstract}
 Human writers often \textit{bookend} their writing with ending sentences that relate back to the beginning sentences in order to compose a satisfying narrative that ``closes the loop.'' Motivated by this observation, we propose \model, a controllable story-generation paradigm that generates narratives by ensuring the first and last sentences are related and then infilling the middle sentences. Our contributions include an initial exploration of how various methods of bookending from Narratology affect language modeling for stories. Automatic and human evaluations indicate \model produces better stories with more narrative closure than current autoregressive models.
\end{abstract}

\section{Introduction}

\textit{Narrative closure} is an important feature of satisfying narratives. \citet{carroll2007narrative} defines narrative closure as ``the phenomenological feeling of finality that is generated when all the questions saliently posed by the narrative are answered.'' Human writers often achieve closure through \textit{bookending} \cite{adamo1995beginnings} (a.k.a circular construction or ring composition) whose minimum criteria is for the ending to relate back to the beginning \cite{novakovich2008fiction, katz2023here}.

Automatic story generation has advanced significantly recently \cite{chaturvedi2016ask, chaturvedi-etal-2017-story, peng-etal-2017-joint, fan2018hierarchical, yao2019plan, fan2019strategies, brahman-chaturvedi-2020-modeling, brahman2020cue, freiknecht2020procedural, castricato2021tell, chowdhury2021everything, vijjini2022towards, yang-etal-2022-re3, huang-etal-2023-affective}. However, these approaches still struggle to generate satisfying and coherent stories with closure \cite{alabdulkarim-etal-2021-automatic, piper2021narrative}. To address this challenge, we propose \textbf{R}elated \textbf{E}ndpoint \textbf{Nar}rative \textbf{Gen}erator (\model)\footnote{Code/resources: \href{https://github.com/adbrei/RENarGen}{https://github.com/adbrei/RENarGen}} to generate closed narratives via bookending with related first and last sentences.

\begin{figure}[t]
    \centering
    \includegraphics[width=0.49\textwidth]{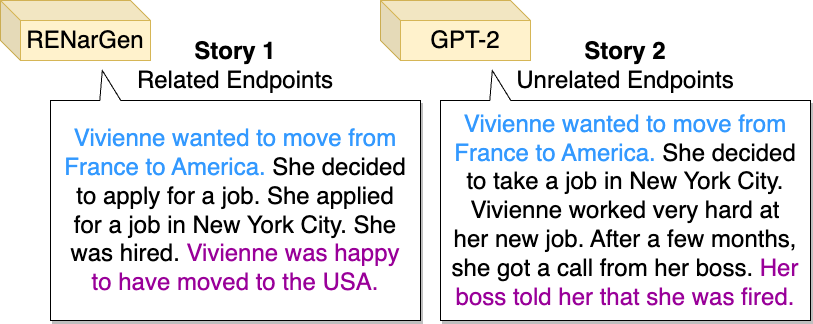}
    \caption{Stories with related \textcolor{startblue}{start} and \textcolor{violet}{stop} sentences (Story 1, generated by \model) provide better narrative closure than stories with unrelated endpoints (Story 2, generated by GPT-2 baseline).}
    \label{fig:two_stories}
\end{figure}

\begin{figure*}[t]
    \centering
    \includegraphics[width=\textwidth]{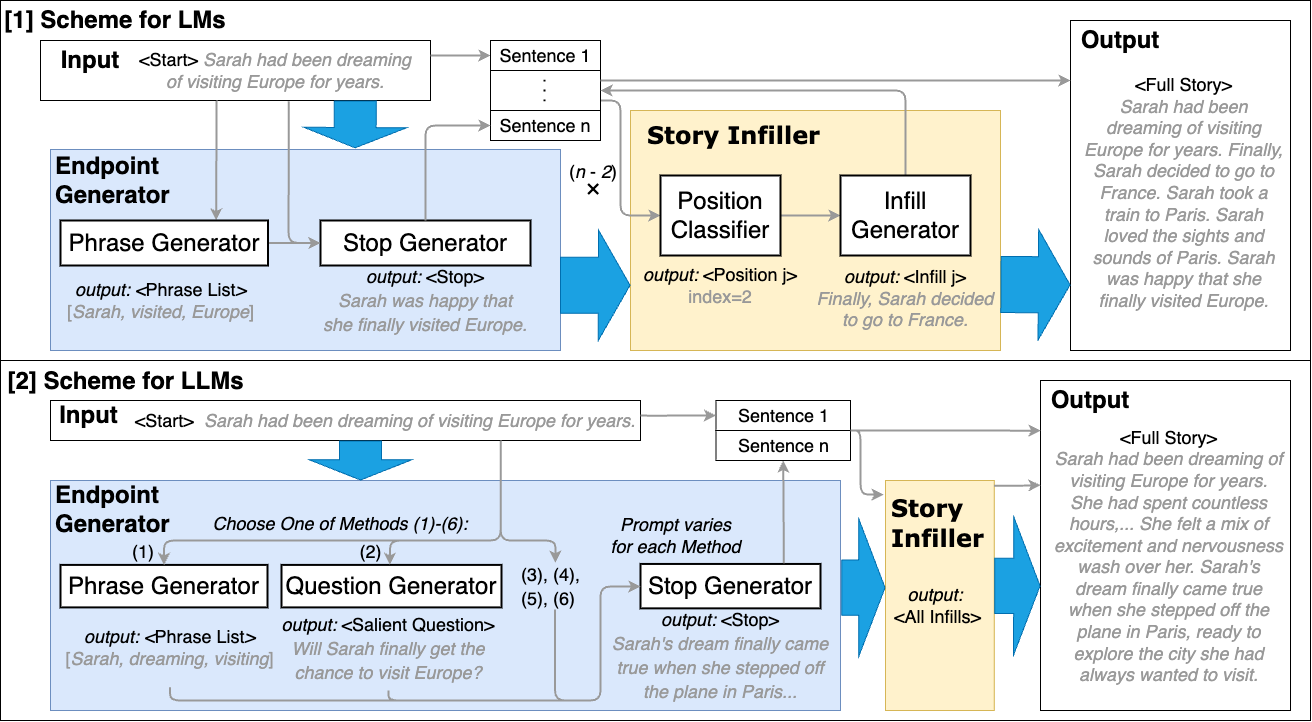}
    \caption{Proposed \model framework. \textbf{Box 1}: Scheme for LMs. Given input start, the Phrase Generator produces a phrase list of relatable words; using this list, the Stop Generator outputs the stop. The Story Infiller infills middle sentences by iteratively determining the next best location for a new sentence and generating a sentence. A sample step-by-step story generation is given in Appendix \ref{sec:gen_example}. \textbf{Box 2}: Scheme for LLMs. Given input start, Endpoint Generater chooses one of six methods to generate the stop. The Story infiller uses the start and stop to generate all infills. After data cleaning, all components are concatenated into the final full story.}
    \label{fig:renargen-framework}
\end{figure*}

We refer to the first sentence as the \textit{start}, the last sentence as the \textit{stop}, and the start/stop sentence pair as \textit{endpoints}. 
Narrative closure can be achieved via \textit{related endpoints}, which may be operationalized with various methods, the most common of which is semantic relatedness. 
Endpoints are semantically related \cite{vmohammad2008measuring,abdalla2021makes} if they resemble each other w.r.t. elements like theme, character, action, place. Figure \ref{fig:two_stories} illustrates this idea with two stories: Story 1 has related endpoints sharing semantic commonalities that complete themes introduced in the start (e.g., protagonist $\rightarrow$ Vivienne, action $\rightarrow$ moving, and place $\rightarrow$ USA); Story 2 has unrelated endpoints with fewer semantic similarities; the stop introduces new themes without satisfactorily fulfilling the initial narrative thought. To a reader, stories like Story 1 are more ``closed'' than stories like Story 2.

\model (Figure \ref{fig:renargen-framework}) is a scheme that produces stories with closure using neural language models (LMs) and large language models (LLMs) by (1) generating related endpoints given the start and (2) infilling middle sentences given left and right contexts. We approach these two challenges differently for LMs versus for LLMs. For the first challenge for LMs, 
we use semantic relatedness to encourage narrative closure.  
We generate a phrase list (salient words/phrases from the start) to emphasize narrative aspects that should be addressed in the stop.
For LLMs, we experiment with different single/multi-prompting methods that address bookending with more sophisticated definitions of relatedness for narrative closure. 
For the second challenge for LMs, we propose an interative infilling method, 
inspired by story-completing techniques described in Narratology \cite{Zemliansky_2020}, that considers both left and right contexts and generates any number of sentences in a reasonable order. While adding sentences left-to-right is a common method of expanding a story, infilling is also a bonafide method: the basic intuition is to find two consecutive sentences between which additional story material is needed. Infilling imitates human writers who add sentences to earlier locations where they determine additional information is necessary \cite{Zemliansky_2020, flower1981cognitive, van2003writing, milligan2017formal, turner2009influence}. Our method is different from previous works using an automatic bidirectional attention strategy \cite{devlin2018bert, ippolito-etal-2019-unsupervised, gu2019insertion, song2019mass, zhu2019text, wang2020narrative, joshi2020spanbert, donahue-etal-2020-enabling} that require the infill to have fixed length, require knowledge of the infill location at the beginning of inference, and/or are not easily iterable. For LLMs, the Story Infiller generates all infills at one time.

Through piece-wise narrative generation, \model offers user interactivity. For example, for LMs the user can control the generated stop sentence by editing the phrase list. 

\model uses both LMs and LLMs because both have their strengths. LMs are more accessible with predictable output format but are less coherent. LLMs produce higher quality generation but are less accessible and require more computing power. See Appendix \ref{sec:lm_llm} for further discussion. 

Automatic and human evaluations indicate \model outperforms baselines with stories that feel more complete. Our contributions are:

\begin{itemize}[itemsep=0pt, topsep=0pt, noitemsep]
    \item We present the first study of how related endpoints affect narrative generation with an early outlook on how the ``good writing practice'' of bookending impacts language modeling. 

    \item We propose \model, a paradigm adaptable to LMs and LLMs and that produces narratives with related endpoint sentences using a novel infilling strategy.

    \item We conduct automatic and human evaluations to show that the stories generated by \model have related endpoints that help with narrative closure and that improve coherence.
\end{itemize}

\section{\model}

Given start sentence, $s_1$, \model generates a story $S=\{s_1,s_2,\dots,s_n\}$ where $n$ is the total number of sentences and the endpoints $s_1$ and $s_n$ are related. \model has two main components: the Endpoint Generator and Story Infiller.\footnote{See Appendix \ref{sec:implementation_details} for additional implementation details.}

\subsection{Endpoint Generator for LMs}

Given start, $s_1$, the Endpoint Generator produces a related stop, $s_n$. This component has a \textit{Phrase Generator} that generates a phrase list of relatable tokens from the start and a \textit{Stop Generator} that generates the stop for the given start incorporating the phrase list.\footnote{We determine that by using a phrase list of related words, the generated stop incorporates aspects from the start, thereby addressing themes and potentially questions raised by the start.} See Figure \ref{fig:renargen-framework} for example generations.

The Phrase Generator is an LM that autoregressively outputs a phrase list, $l=[t^1,\dots, t^r]$ where each $t^i$ is a token with the potential to relate $s_1$ and a future $s_n$.  
For our experiments, we use GPT-2 \cite{radford2019language} fine-tuned on pairs of start sentences, $s_1$, and their corresponding phrase lists, $l$, extracted from our dataset (Sec. \ref{section:data}). For extracting phrase lists, we measure similarity of each start token embedding with each stop token embedding via cosine similarity of BERT (uncased) \cite{devlin2018bert} embeddings and extract stop tokens with similarity greater than threshold, $\gamma$.\footnote{For this task, we use $\gamma=0.7$, a high threshold to ensure the phrase list contains only the most relevant related tokens.} 

The Stop Generator autoregressively accepts the concatenation of $s_1$ and $l$ and outputs $s_n$. We use GPT-2 fine-tuned on triples of starts, extracted phrase lists, and stops of stories from our dataset.

\subsection{Story Infiller for LMs}
The Story Infiller generates the sentences between the endpoints, $\{s_2,s_3,\dots,s_{n-1}\}$. It does not infill the sentences in a left-to-right manner and instead dynamically decides where to infill a sentence by determining where context is missing. The Story Infiller consists of two models: a \textit{Position Classifier} and an \textit{Infill Generator}. 

The Position Classifier analyzes all positions between consecutive sentences in the story so far and decides the infilling position, $i \in \{2, 3, \ldots, n-1\}$. The $i$ is the index that needs the most information for the story to sound coherent. We fine-tune BERT (uncased) to predict if the story is missing a sentence at a given position.
We construct positive examples by randomly masking $1$-$3$ sentences per story in our dataset. We construct negative samples by inserting one mask token where the story is not missing any sentences.
During inference, the model considers all possible infilling positions in an incomplete story, and the position with maximum probability is selected as the next infill location.

The Infill Generator generates the missing sentence, $s_{i}$. We fine-tune GPT-2 on samples with $s_1,\dots, s_{i-1}\ \langle mask \rangle\ s_{i+1},\dots, s_{n}\ \langle sep\rangle\ s_{i}$ to generate $s_{i}$. We insert the generated sentence, $s_{i}$ into the story, $s$, and repeat the infilling process until $n-2$ sentences are infilled. We note $n$ is a flexible threshold specified by the user. 

Through this process, the Story Infiller (1) does not depend on a specific location for infill, (2) considers both left and right contexts, and (3) considers all sentences in the context, where $n$ may be an arbitrary number of sentences set by the user.
Appendix \ref{sec:long_gen_lm} shows examples of stories of varying $n$-sentence lengths generated by \model.

\subsection{\model for LLMs}

% \begin{table*}[t]
%     \centering
%     \footnotesize
%     \begin{tabular}{l|c|c|c||c|c}  
%         Models      & Lexical Overlap & Cosine Sim. & Syntax Sim. & Distinct n-grams & BLEU\\
%         \hline
%         GPT-2       & 0.183$\pm  $0.123    &0.458$\pm$0.143     &0.533$\pm$0.112  &0.42016 &3.14$\pm0.15$ \\
%         \model-LM  &\textbf{0.329}$\pm$0.136  &\textbf{0.653}$\pm$0.121  &\textbf{0.594}$\pm$0.110 & \textbf{0.52416} & \textbf{3.35}$\pm0.15$\\
        
%         \hline
        
%         Llama-7b-chat       & 0.494$\pm$0.093 & 0.772$\pm$0.074 & 0.207$\pm$0.048 & 0.762 & 1.613$\pm$1.432 \\
%         \model-LLM-7b (1)   & \textbf{0.509}$\pm$0.081 & \textbf{0.829}$\pm$0.052 & \textbf{0.214}$\pm$0.026 & \textbf{0.773} & \textbf{1.622}$\pm$0.936 \\

%         \hline
%         Llama2-70b-chat     & 0.476$\pm$0.071 & 0.772$\pm$0.066 & 0.199$\pm$0.005 & 0.783 & 2.140$\pm$1.407 \\
%         \model-LLM-70b (1)  & \textbf{0.522}$\pm$0.069 & \textbf{0.842}$\pm$0.040 & \textbf{0.199}$\pm$0.005 & \textbf{0.798} & \textbf{2.415}$\pm$1.299 
%     \end{tabular}
    
%     \caption{Automatic evaluation of endpoint relatedness (first 3 cols) and overall quality (last 2 cols). Bold text indicates statistical significance, $p<0.05$ \cite{dror2018hitchhiker}. Results for LLMs were conducted on a subset of the data for resource and computational cost considerations. \model generates more coherent and closed stories.}
%     \label{tab:autoresults}
% \end{table*}

\begin{table*}[t]
    \centering
    \footnotesize
    \begin{tabular}{l|c|c|c||c|c}  
        Models      & Lexical Overlap & Cosine Sim. & Syntax Sim. & Distinct n-grams & BLEU\\
        \hline
        \model-LM   &\textbf{0.329}$\pm$0.136  &\textbf{0.653}$\pm$0.121  &\textbf{0.594}$\pm$0.110 & \textbf{0.524} & \textbf{3.35}$\pm0.15$\\
        \hspace{.25cm}\textit{w/out PG} & 0.298$\pm$0.124 & 0.622$\pm$0.122 & 0.595$\pm$0.111 & -- & -- \\
        \hspace{.25cm}\textit{w/out PC} & -- & -- & -- & 0.4346 & 2.93$\pm$0.16\\
        GPT-2 Baseline  & 0.183$\pm  $0.123    &0.458$\pm$0.143     &0.533$\pm$0.112  &0.420 &3.14$\pm0.15$ \\
        
        \hline
        
        \model-LLM-7b (1)   & 0.509$\pm$0.081 & 0.829$\pm$0.052 & 0.214$\pm$0.026 & \textbf{0.773} & 1.622$\pm$0.936 \\
        \model-LLM-7b (2)   & 0.562$\pm$0.084 & 0.844$\pm$0.055 & 0.208$\pm$0.033 & 0.761 & \textbf{1.661$\pm$0.971} \\
        \model-LLM-7b (3)   & 0.572$\pm$0.091 & 0.847$\pm$0.055 & 0.212$\pm$0.054 & 0.758 & 1.572$\pm$0.791 \\
        \model-LLM-7b (4)   & \textbf{0.589}$\pm$0.103 & \textbf{0.854}$\pm$0.059 & \textbf{0.252}$\pm$0.093 & 0.748 & 1.579$\pm$0.897 \\
        \model-LLM-7b (5)   & 0.520$\pm$0.096 & 0.795$\pm$0.087 & 0.203$\pm$0.022 & 0.767 & 1.578$\pm$0.942 \\
        \model-LLM-7b (6)   & 0.565$\pm$0.077 & 0.844$\pm$0.052 & 0.205$\pm$0.023 & 0.770 & 1.766$\pm$1.252 \\
        \hspace{.25cm}\textit{w/out EG \& SI} & 0.491$\pm$0.133 & 0.749$\pm$0.091 & 0.217$\pm$0.055 & 0.763 & 1.649$\pm$1.160 \\
        Llama-7b-chat Baseline & 0.494$\pm$0.093 & 0.772$\pm$0.074 & 0.207$\pm$0.048 & 0.762 & 1.613$\pm$1.432 \\

        \hline
        \model-LLM-70b (1)  & 0.522$\pm$0.069 & 0.842$\pm$0.040 & 0.199$\pm$0.005 & \textbf{0.798} & \textbf{2.415}$\pm$1.299 \\
        \model-LLM-70b (2)  & 0.526$\pm$0.071 & 0.844$\pm$0.053 & 0.195$\pm$0.015 & 0.791 & 1.797$\pm$1.526 \\
        \model-LLM-70b (3)  & \textbf{0.594}$\pm$0.0753 & \textbf{0.870}$\pm$0.045 & 0.199$\pm$0.005 & 0.787 & 1.576$\pm$0.916 \\
        \model-LLM-70b (4)  & 0.523$\pm$0.0872 & 0.061$\pm$0.566 & 0.192$\pm$0.022 & 0.783 & 1.877$\pm$1.476 \\
        \model-LLM-70b (5)  & 0.512$\pm$0.0844 & 0.064$\pm$0.576 & 0.194$\pm$0.017 & 0.783 & 2.239$\pm$2.102 \\
        \model-LLM-70b (6)  & 0.512$\pm$0.0935 & 0.083$\pm$0.398 & 0.192$\pm$0.019 & 0.782 & 2.244$\pm$1.874 \\
        \hspace{.25cm}\textit{w/out EG \& SI} & 0.526$\pm$0.085 & 0.805$\pm$0.082 & 0.192$\pm$0.020 & 0.795 & 2.351$\pm$1.385 \\
        Llama2-70b-chat Baseline & 0.476$\pm$0.071 & 0.772$\pm$0.066 & 0.199$\pm$0.005 & 0.783 & 2.140$\pm$1.407 \\
    \end{tabular}
    
    \caption{Automatic evaluation of endpoint relatedness (first 3 cols) and overall quality (last 2 cols). Indented models are ablation studies. Bold text indicates statistical significance, $p<0.05$ \cite{dror2018hitchhiker}. Results for LLMs were conducted on a subset of the data for resource and computational cost considerations. \model generates more coherent and closed stories.}
    \label{tab:autoresults}
\end{table*}
\begin{table}
    \footnotesize
    \begin{tabular}{l|c|c|c|c}
                    & Rel. & Clos. & Coh. & Pref. \\
         \hline
         \model-LM  &  \textbf{0.63} &  \textbf{0.47} & \textbf{0.62} & \textbf{0.66}\\
         GPT-2      &  0.20 &  0.18 & 0.21 & 0.21\\
         Tie        &  0.17 &  0.35 & 0.17 & 0.13\\
         \hline
         \model-LLM-7b & \textbf{0.58} & \textbf{0.56} & \textbf{0.55} & \textbf{0.56} \\
         Llama-7b      & 0.39 & 0.43 & 0.41 & 0.43\\
         Tie           & 0.03 & 0.01 & 0.04 & 0.01\\
         \hline
         \model-LLM-70b & \textbf{0.80} & \textbf{0.56} & \textbf{0.56} & \textbf{0.56} \\
         Llama-70b       & 0.20 & 0.44 & 0.44 & 0.44\\
         Tie            & 0.0 & 0.0 & 0.0 & 0.0
    \end{tabular}
    \caption{Human evaluation of \model vs baselines, showing humans prefer \model-generated stories. Bold text indicates statistical significance, $p<0.05$.}
    \label{tab:results1}
\end{table}

Shown in Box 2 of Figure \ref{fig:renargen-framework}, \model for LLMs also uses an Endpoint Generator and Story Infiller. The Endpoint Generator prompts a pre-trained LLM using one of a set of methods specified by the user during inference to generate endpoints based on various definitions of bookending for narrative closure from narratology theory:

\begin{enumerate}[label=(\arabic*), itemsep=0pt, topsep=0pt]
    \item Prompt for a phrase list $p$ from the start and generate a corresponding stop (parallels the structure of \model-LM);
    
    \item Prompt for a ``related'' stop given the start and the LLM's pre-trained knowledge of sentence relatedness;
    
    \item Prompt for the salient narrative question introduced by the start and generate a stop that answers the question. Addresses the erotetic definition of story closure \cite{carroll2007narrative} by concluding the salient narrative question;
    
    \item Generate a stop with the same character, related action, and/or location as the start. Seeks stricter control for specific narrative elements and follows the ``matching ending'' technique \cite{novakovich2008fiction} for narrative closure;
    
    \item Generate a stop that entails the start. Hence, the truth of the stop logically leads to the truth of the start, resulting in semantically close endpoint sentences;
    
    \item Generate a stop entailed by the start. Hence, the truth of the start logically leads to the truth of the stop, resulting in semantically close endpoint sentences.
\end{enumerate}

See Appendix \ref{sec:llm_prompts} for additional prompting details.

The Story Infiller receives the start and generated stop and infills all sentences (an arbitrary number or a specified $n$) in a left-to-right manner. For our experiments, we generate 5-sentence stories with pre-trained Llama2 models \cite{touvron2023llama}.
% with temp $=0.9$, top\_p $=0.9$. 
Examples of longer generations are given in Appendix \ref{sec:long_gen_llm}

\section{Empirical Evaluation}

\subsection{Dataset}
\label{section:data}
We use the ROCStories corpus \cite{mostafazadeh-etal-2016-corpus}, a collection of 5-sentence human-written stories. For \model for LMs, we combine Spring 2016 and Winter 2017 sets and obtain 98,161 stories which are split 80:20 for training and validation. For evaluation, we use the 3742 stories from Cloze Spring 2016.

\subsection{Automatic Evaluation}
\label{sec:automatic_evaluation_main}

For LMs, we compare \model with a baseline GPT-2 fine-tuned on all training samples in the ROCStories corpus; at runtime, given the start, the baseline generates a corresponding five-sentence story in a left-to-right manner. 
For LLMs, the baselines are Llama2-7b and Llama2-70b, where prompts do not specify endpoint relatedness. 
We evaluate endpoint relatedness and overall quality of generated stories. Table \ref{tab:autoresults} shows the results.

Evaluating endpoint relatedness (or narrative closure) is challenging. In this work, we quantify it automatically with five metrics. 
We compute \textit{Lexical overlap} via Dice Coefficient \cite{Saad2013dice}, \textit{Cosine similarity} with Sentence-BERT embeddings \cite{reimers2019sentence}\footnote{Endpoint relatedness is measured via cosine similiarity of start and stop sentence embeddings generated by SentenceBERT fine-tuned on \textit{STR-2022} \cite{abdalla2021makes}, a dataset of 5,500 English sentence pairs with relatedness scores.}, and \textit{Syntax similarity} with FastKASSAM \cite{chen-etal-2023-fastkassim,boghrati2018conversation} that uses a label-based tree kernel.  For overall quality, we compute the average of all \textit{distinct n-grams} ($n=\{1\dots5\}$) for measuring repetition and lexical creativity, and \textit{comparison against reference stories} with BLEU score \cite{papineni2002bleu, post2018call}. For all of these measures, a higher value is better.

From the endpoint relatedness scores, we see \model is capable of generating more related endpoints than the baselines. From the overall quality scores, we see \model generates more coherent stories with more diverse content.

We conduct ablations to test the importance of various components of \model. 
For \model-LM, the ablation experiments remove (1) the Phrase Generator by generating the stop directly from the start and (2) the Position Classifier by adding each new sentence to a randomized position. For \model-LLM, the ablation removes the Story Infiller by simultaneously generating the entire story from left-to-right with a specified stop related to the start.
Our experiments indicate the Phrase Generator and Position Classifier for LMs and the Endpoint Generator and Story Infiller for LLMs are important.

\subsection{Human Evaluation of Quality}
\label{sec:amt_description}
We conducted human evaluations on the Amazon Mechanical Turk (AMT) platform.  We randomly sampled generated stories and performed pairwise comparisons between \model and corresponding baselines. Presentation order of the stories was randomized. Evaluators were asked to select the story with the better related endpoints, sense of closure, coherency, and overall quality. The evaluators were asked to not consider other criteria while evaluating on a specific criterion, except when judging overall quality. Sample story pairs and the instructions are shown in Appendix \ref{sec:mod_v_gpt} and \ref{sec:amt_eval1}. We limited the task to USA-based master workers with 98\% approval rates and more than $5000$ approved HITs.  We evaluated $100$ story pairs per comparison. Results (see Table \ref{tab:results1}) show evaluators preferred \model stories across all criteria. This indicates \model can produce coherent narratives that are better at providing a sense of closure to the human reader. 

\subsection{Human Evaluation of Interactivity} 

We conducted a human evaluation of interactivity with \model-LM. We asked 8 in-house testers (native English speakers with minimum higher education degree of Bachelor's) to edit phrase lists generated by the Phrase Generator on $50$ unique starts. Evaluators had unlimited editing attempts per input. At the end of each interaction, they answered (1) whether or not they could generate better stories than the initial \model stories via interactivity, and (2) how useful they found the feature of editing phrase lists. For the majority of the interactions ($80\%$), users found that the ability to control the phrase list enabled them to generate better stories than the automatically generated \model stories, and $62.5\%$ users ranked the usefulness of interactivity as 4 on a $0$-$5$ scale (5 being most useful). On average, users tried $3$ unique phrase lists per start. These results indicate the phrase list is important for story generation and users enjoy having the ability to control this aspect.

\section{Conclusion} 
We present \model to automatically generate stories with related endpoints via various methods of bookending. \model for LMs uses semantic relatedness and \model for LLMs uses several forms of semantic relatedness, erotetic closure, ``matching ending,'' or entailment to guide the generation of related endpoints.
We empirically demonstrate \model produces more closed, satisfying, and coherent narratives than corresponding baselines. 
We also show that users find \model-LM's element of controllability useful. Through our experiments and corresponding human evaluations for pair-wise preference, we further demonstrate the applicability of narratology theory for improved automatic generations and the importance of narrative closure for satisfying narratives.

\section{Acknowledgement}
The authors are thankful to the anonymous reviewers, and to Somnath Basu Roy Chowdhury, Haoyuan Li, and Anvesh Rao Vijjini for their constructive comments. We thank Amartya Banerjee, Nathan Brei, ML Brei, William Brei, Denali Dahl, Katharine Henry, Benjamin Linford, Vaidehi Patil, Marlus Pedrosa, Simantika Roy, Lindsey Whitlow for their feedback on evaluating interactivity and/or using \model.

\section{Limitations}

Since the generative model components of \model have been fine-tuned solely on datasets of stories written in English, \model can only generate text in English. For similar reasons, due to its training data, it is also limited to generating story narratives.

\section{Ethics Statement} 

The GPT-2 components of \model are fine-tuned on the ROCStories corpus, a dataset which has been shown to have gender bias \cite{huang2021uncovering}. As such our system might replicate or amplify this bias and other potential biases in the training dataset.

%%%%%%%%%%%%%%%%%%%%%%%%%%

\bibliography{custom}

\appendix

\section{Example Generation with \model}
\label{sec:gen_example}

    We demonstrate the steps \model takes to generate a story by giving an example. First, a start sentence from the testing set is given as input. Then a corresponding stop sentence is generated. For each $n-2$ middle sentences (where $n$ is the total number of sentences to be generated), the Story Infiller iteratively chooses a location that is missing a sentence and produces a sentence to fit it. The output is the complete story with the start, all infilled sentences, and the stop. In the below steps, we show how \model takes the input and generates a five-sentence story.\\

    \textbf{Input}:
    A husband and his wife are looking for a new home.
    
    \noindent\hdashrule{\linewidth}{1pt}{1pt}
    \textbf{Stop generation}:\\
    They are excited to finally have a home!
    
    \noindent\hdashrule{\linewidth}{1pt}{1pt}
    \textbf{Story infiller}:
    \begin{center}
        \textit{Iteration 1}
    \end{center}
    A husband and his wife are looking for a new home. \textit{$\langle$They have been looking for months.$\rangle$} They are excited to finally have a home!
    \begin{center}
        \textit{Iteration 2}
    \end{center}
    A husband and his wife are looking for a new home. They have been looking for months. \textit{$\langle$They finally found one in their area.$\rangle$} They are excited to finally have a home!
    \begin{center}
        \textit{Iteration 3}
    \end{center}
    A husband and his wife are looking for a new home. They have been looking for months. \textit{$\langle$Finally they have found the perfect place.$\rangle$} They finally found one in their area. They are excited to finally have a home!
    
    \noindent\hdashrule{\linewidth}{1pt}{1pt}
    \textbf{Output}:\\
    A husband and his wife are looking for a new home. They have been looking for months. Finally they have found the perfect place. They finally found one in their area. They are excited to finally have a home!

\section{Use of LMs vs. LLMs}
\subsection{Advantages/Disadvantages}
\label{sec:lm_llm}

LMs are advantageous over LLMs because they are smaller and more accessible models. As a result of their fine-tuning, their output is more consistent, resulting in less noise. Data cleaning between components is more easily standardized.

LLMs are advantageous over LMs because their larger architecture makes it possible to generate more sophisticated stories with greater coherence and interesting sentence structure. LLMs also provide additional explainability regarding why output is relevant. For example, the output often includes additional description regarding how it satisfies the prompt criteria. We make use of explainability in \model-LLM Experiment (2): while prompting for a stop that answers questions introduced in the start, we request the model first explain salient questions raised in the start. 

However, LLMs tend to generate less consistent output with much noisy chatter. As a consequence, the quality tends to vary wildly generation-to-generation and prompt-to-prompt. The intermediate output is more difficult to clean on a large scale, since there is much variance. Additionally, we note that Llama2-7b does not complete many stories whose start contains a negative sentiment due to the model's severe sensitivity to potentially harmful content. 

\subsection{Experimental Comparison}

Experiments are conducted on Linux and macOS. For \model-LM we use versions of GPT-2 (1.5 billion parameters) and BERT (110 million parameters) under MIT and Apache licenses respectively. For \model-LLM we use Llama2-7b-chat (7 billion parameters) and Llama2-70b-chat (65 billion parameters) under the Llama2 license. Models are used for purposes consistent with their intended use. All experiments use a varying number of GPU hours depending on the length of each story. For a maximum length of 512 tokens, each story takes less than a minute to generate.

\section{Implementation Details}
\label{sec:implementation_details}

Code is primarily implemented in Python. Existing packages and libraries used for preprocessing, fine-tuning, and running \model include: datasets, numpy, ollama, pandas, random, re, sentence\_transformers, sklearn.model\_selection, torch, tqdm, transformers. Evaluation metrics are specified in Section \ref{sec:automatic_evaluation_main}. For BLEU scores, we use sacreBLEU \cite{post2018call}. For all packages, default parameters are used unless otherwise specified.

\section{Example \model stories of different lengths}
\label{sec:long_gen}
\subsection{\model with LMs}
\label{sec:long_gen_lm}

    We include samples of generated stories of different lengths. In this way, we demonstrate the capability of \model to generate stories of arbitrary lengths $n=\{8,10,15,10,25\}$ using our Story Infiller. Below we show five example stories generated by \model:\\
    
    \noindent\textbf{Example 1}: \textit{Number of sentences = 8}
    
    \noindent It was  my final performance in marching band. It was my first marching competition. I had been practicing for weeks. When I got on stage, I was nervous. I took the stage and began to sing. When the bell rang, everyone cheered. The band played very well. I finished the band and was very proud of myself.\\
        
    \noindent\textbf{Example 2}: \textit{Number of sentences = 10}
    
    \noindent At school, Mary received an assignment to write an essay about pandas. She was very nervous about the assignment. She did not know what to say. She decided to give it a try. She tried to think of a good answer. Her teacher asked her what she thought. The teacher told her that it was a bad idea. Mary was afraid she would get an F. When she got to class, her teacher gave her a big hug. Mary wrote the essay and got an A on it.\\
        
    \noindent\textbf{Example 3}: \textit{Number of sentences = 15}
    
    \noindent The delivery man handed a package to me. When I opened it, there was a large box. Inside the box was an envelope. It was addressed to my wife. The package was wrapped in wrapping paper. The envelope contained several gifts. Inside of the envelope was several envelopes. My wife was very excited. The box contained gifts for me to give her. Inside it was the address for the gift. It had a note inside. I took it to the post office. Inside, I found a gift certificate. Inside were several presents. I delivered the package.\\
        
    \noindent\textbf{Example 4}: \textit{Number of sentences = 20}
    
    \noindent Helen lived in the state of Washington. She loved the outdoors. She had a lot of outdoor activities. One day, a hurricane came through the area. Helen decided to move to Florida. The weather was terrible. Helen's house was destroyed. Helen lost power. Helen had to evacuate her house. A hurricane was headed towards her home. Hurricane season was approaching. It made it hard for Helen to stay indoors. It was very stormy. There was a huge storm coming through. It rained heavily. It had tornadoes and tornados. A big storm was coming. The hurricane destroyed many homes. Helen was devastated. Now, Helen lives in Washington, DC.\\
        
    \noindent\textbf{Example 5}: \textit{Number of sentences = 25}
    
    \noindent Oliver needed a new radio. His old one was broken. He could not afford it. Oliver decided to look online. He found a radio store. He couldn't find any good ones. He decided he needed to save money. He needed one for his house. He didn't have enough money for it though. He asked his friends for advice. His friend told him he could get one at a cheaper price. So he looked around. He saw a great deal on a good radio there. Oliver was very upset. He was sad. Oliver knew his old radio was too expensive. He knew he would have to find a replacement. Oliver went shopping. He did not know what to do. He had to go to a store to buy one. He searched everywhere. Oliver looked online for radio parts. He looked at many radio stores. Finally he found one he liked. He went to the store and bought the radio he wanted.
    
\subsection{\model with LLMs}
\label{sec:long_gen_llm}

Below are samples from \model-LLM (2):\\

\noindent\textbf{\model-7b Example}:

Ron started his new job as a landscaper today with high hopes and expectations. He had always been passionate about nature and the outdoors, and he was thrilled to have the opportunity to turn that passion into a career. As he arrived at the job site, he was greeted by his supervisor, who showed him around the property and introduced him to the other members of the team. Ron was eager to get started and prove himself, but as he began to work, he quickly realized that the job was much more physically demanding than he had anticipated. Despite feeling exhausted and overwhelmed, Ron pushed through and worked tirelessly throughout the day, determined to make a good impression on his first day.

As the days went by, Ron continued to work hard and prove himself as a valuable member of the team. He learned the ins and outs of landscaping and gardening, and he took great pride in his work. However, despite his dedication and effort, Ron was presented with a grueling work schedule that left him feeling drained and demotivated. Despite this, he refused to give up and continued to push himself to the limit, determined to succeed in his new career.

As the weeks went by, Ron's hard work began to pay off, and he started to see the fruits of his labor. The garden he had helped create was thriving, and the property was looking better than ever. Ron felt a sense of pride and accomplishment, knowing that he had played a role in making it happen.

Despite the challenges he faced, Ron remained determined to succeed in his new job as a landscaper. He continued to work hard and prove himself, and he was grateful for the opportunity to turn his passion into a career. In the end, Ron's excitement to start his new job had been dampened by the grueling work schedule, but he was determined to make it work and prove himself as a valuable member of the team.\\

\noindent\textbf{\model-70b Example}:

Early every morning the hummingbirds visit my porch.
The tiny birds flit from flower to flower, their rapid wings beating a blurry haze around them.
They sip nectar from the blooms, their long, slender beaks inserting into the petals with precision.
With a twinkle of green and blue, a hummingbird hovers in front of me, its gaze seeming to meet mine.
I watch, mesmerized, as it darts and weaves, its wings a whir of motion.
The hummingbirds are a delightful distraction from the stresses of my day, their antics a reminder to slow down and appreciate the beauty in the world.
As I sit on my porch, sipping my morning coffee, the hummingbirds' visits become a familiar comfort, a signal that all is well in my corner of the universe.
Their daily appearances are a reminder of the cyclical nature of life, a promise that each day brings new opportunities for wonder and joy.
And so, I savor each moment with the hummingbirds, knowing that their presence is a gift, a fleeting glimpse of magic in an often-mundane world.
As the sun sets, the hummingbirds return to their nests, their iridescent feathers glistening in the fading light, and I am left to ponder the magic of their daily visits to my porch.

\section{Prompts for LLM Experiments}
\label{sec:llm_prompts}

\noindent \noindent We provide the prompts used for LLM experiments. We control narrative length by specifying the number of sentences to be generated. Word-choice variation across models is minimal.

\subsection{System prompts}
% \noindent\textbf{System prompts:} 

\noindent\textbf{Llama2-7b:} ``You are a talented writer. Generate sentences for a well-written narrative. If you have ethical concerns, resolve them in the story.''\\

\noindent\textbf{Llama2-70b:} ``You are a talented writer. For each prompt, only generate the sentences for a well-written narrative.''

\subsection{Endpoint Generator}

\noindent (1.1) ``Here is the first sentence of a narrative: $\langle start\rangle$.
What are the most salient words or phrases? Give me a list, where each item is separated by a comma.'' \\

\noindent (1.2) ``Here is the first sentence and its salient words/phrases: $\langle start, l\rangle$. Using this first sentence and the list of salient words/phrases, give one related closing sentence'' \\

\noindent (2) ``Here is the first sentence of a narrative: $\langle start\rangle$. Please give me a closing sentence which is related to the first sentence.'' \\

\noindent (3.1) ``Here is the first sentence of a narrative: $\langle start\rangle$. What is the most salient question to propel the narrative forward?'' \\

\noindent (3.2) ``Here is the first sentence and relevant question for a narrative: $\langle start,$ $3.1\ output\rangle$. Give me ONE closing sentence that answers the most salient question without introducing new questions.'' \\

\noindent (4) ``Here is the first sentence of a narrative: $\langle start\rangle$. Please give me a closing sentence that has the same character and/or same related action and/or location.'' \\

\noindent (5) ``Here is the first sentence of a narrative: $\langle start\rangle$. Please give me a closing sentence that entails the first sentence.'' \\

\noindent (6) ``Here is the first sentence of a narrative: $\langle start\rangle$. Please give me a closing sentence that is the entailment of the first sentence.''

\subsection{Story Infiller}

\noindent (1-5) ``Here is the first sentence of a narrative: $\langle start\rangle$ and here is the last sentence: $\langle 1$-$5\ output\rangle$. What happens between these sentences? Please give me THREE consecutive intermediate sentences.''

\subsection{Long generation} 

\noindent``Here is the first sentence of a narrative: $\langle start\rangle$ and here is the last sentence: $\langle stop\rangle$. What happens between these sentences? Please give the complete story.''

\subsection{Baseline} 

\noindent``Complete the story in FOUR sentences: $\langle start\rangle$.'' 

\subsection{Ablation w/out EG \& SI} 

\noindent``Here is the first sentence of a narrative: $\langle start\rangle$. Please give me the next FOUR sentences. Make sure that the last sentence is related to the first sentence.''

\section{Examples: \model vs GPT-2}
\label{sec:mod_v_gpt}

    We show three example story generations from both \model and GPT-2 to demonstrate the effect of related endpoints on overall narrative closure.
    
    \noindent\rule{\linewidth}{0.5pt}
    
    \noindent\textbf{\model}: Daniel likes to write for fun. He decided to start writing for school. He is very good at it. He gets a lot of feedback. Daniel is happy that he is writing.\\

    \noindent \textbf{GPT-2}: Daniel likes to write for fun. One day, he decided to take a trip to the beach. He went out on the sand and saw a beautiful sunset. Daniel was so happy to see the sunset! Daniel went back home and enjoyed his trip.
        
    \noindent\rule{\linewidth}{0.5pt} 
    
    \noindent\textbf{\model}: Cora was nervous. She went to the doctor. The doctor gave her some medicine. She felt better. Cora was glad she was not nervous anymore.\\

    \noindent \textbf{GPT-2}: Cora was nervous. She didn't know what to do. But she decided to go to the store and buy some candy. When she got home, she opened the box. It was all candy!
        
    \noindent\rule{\linewidth}{0.5pt}

    \noindent\textbf{\model}: Neil had been journeying through Asia. He had never been on a plane before. He boarded the plane. He took a seat in the back seat. Neil was so happy to be on his way to Asia!\\
    
    \noindent \textbf{GPT-2}: Neil had been journeying through Asia. He had never been on a plane before, but he was excited to see the sights. When he got to the airport, he realized he had forgotten his ticket. Neil had to wait in line for an hour for his flight back home. Luckily, Neil was able to board the plane safely.

\section{Instructions for AMT Human Eval}
\label{sec:amt_eval1}
    
    \noindent In this task, we provide two sample stories (Story A and Story B) generated by two artificial intelligence systems. We ask you to compare Story A and Story B and answer the following questions:   
    \begin{enumerate}
        \item \textbf{Consider only the first and last sentences of each story. Which of these two stories has the most related first and last sentences?}

        Sentences are related if they have close meanings via similar semantics (matching words or meanings) and/or syntax (similar sentence structure). For example the first sentence, “Julian ascended the staircase.”, and the last sentence, “Triumphant, Julian descended the staircase”, are related because of semantic similarities (the protagonist is Julian, the action corresponds to vertical movement, and the setting is a staircase) as well as syntactic similarities.
        
        \item \textbf{Overall, which story gives a better sense of closure?}

        After reading each story in its entirety, which one ends more satisfactorily with respect to its beginning? Which story “closes the loop?”
        
        \item \textbf{Which story is more coherent?}

        A coherent story has good flow with a logical structure, smooth transitions, and a unified theme. The story should be relatively easy to read and understand.
        
        \item \textbf{Considering both coherence and closure, which of the two is a better story?}

        Oftentimes readers show different preferences for different stories. Here, we ask you to use your best judgment and select the story that is most satisfying to you considering all the criteria mentioned above.
        
    \end{enumerate}

    When judging on a certain criterion, please do not consider any other criteria. 

    \textbf{Although we provide the “similar” option, and sometimes neither of the stories are perfect, we strongly encourage you to choose a better one from those two, unless they are indeed similar.}

% \begin{table*}[t]
%     \centering
%     \footnotesize
%     \begin{tabular}{l|ccc||cc}  
%     \toprule
%         Models      & Lexical Overlap & Cosine Sim. & Syntax Sim. & Distinct n-grams & BLEU\\\midrule
%         \textit{GPT-2} &&&&&\\
%         Baseline       & 0.183$\pm  $0.123    &0.458$\pm$0.143     &0.533$\pm$0.112  &0.42016 &3.14$\pm0.15$ \\
%         \model-LM   &\textbf{0.329}$\pm$0.136  &\textbf{0.653}$\pm$0.121  &\textbf{0.594}$\pm$0.110 & \textbf{0.52416} & \textbf{3.35}$\pm0.15$\\\midrule
%         \textit{Llama-7b-chat} &&&&&\\
%         Baseline       & 0.494$\pm$0.093 & 0.772$\pm$0.074 & 0.207$\pm$0.048 & 0.762 & 1.613$\pm$1.432 \\
%         \model-LLM (1)   & 0.509$\pm$0.081 & 0.829$\pm$0.052 & 0.214$\pm$0.026 & \textbf{0.773} & 1.622$\pm$0.936 \\\midrule

%         \textit{Llama2-70b-chat} &&&&&\\
        
%         Baseline     & 0.476$\pm$0.071 & 0.772$\pm$0.066 & 0.199$\pm$0.005 & 0.783 & 2.140$\pm$1.407 \\
%         \model-LLM (1)  & 0.522$\pm$0.069 & 0.842$\pm$0.040 & 0.199$\pm$0.005 & \textbf{0.798} & \textbf{2.415}$\pm$1.299 \\
%     \bottomrule
%     \end{tabular}
    
%     \caption{Automatic evaluation of endpoint relatedness (first 3 cols) and overall quality (last 2 cols). Indented models are ablation studies. Bold text indicates statistical significance, $p<0.05$ \cite{dror2018hitchhiker}. Results for LLMs were conducted on a subset of the data for resource and computational cost considerations. \model generates more coherent and closed stories.}
%     \label{tab:autoresults_complete}
% \end{table*}

\end{document}